\pdfoutput=1

\documentclass[11pt]{article}
\usepackage{amsmath} 

\usepackage{acl}
\usepackage{geometry}
\usepackage{booktabs}
\usepackage{graphicx}
\usepackage{float}
\usepackage{multirow}
\usepackage{times}
\usepackage{latexsym}
\usepackage{soul}
\usepackage[colorinlistoftodos]{todonotes}
\usepackage{xcolor}

\usepackage[T1]{fontenc}

\usepackage[utf8]{inputenc}

\usepackage{microtype}

\usepackage{inconsolata}

\usepackage{array}
\usepackage{tabularx}
\usepackage{multirow}

\usepackage{times}
\usepackage{latexsym}
\usepackage{bm}
\usepackage{graphicx}
\usepackage{chngcntr}
\usepackage[T1]{fontenc}

\usepackage[utf8]{inputenc}

\usepackage{microtype}

\usepackage{inconsolata}

\usepackage{caption}
\usepackage{subcaption}

\title{Unlocking Parameter-Efficient Fine-Tuning for \\Low-Resource Language Translation}

\author{Tong Su\textsuperscript{1} \hspace{0.4cm} Xin Peng\textsuperscript{1} \hspace{0.4cm} Sarubi Thillainathan\textsuperscript{2} \\
\textbf{David Guzmán\textsuperscript{1} \hspace{0.4cm} Surangika Ranathunga\textsuperscript{3} \hspace{0.4cm} En-Shiun Annie Lee\textsuperscript{1,4}} \\
\textsuperscript{1}University of Toronto \hspace{0.4cm} \textsuperscript{2}Saarland University \hspace{0.4cm}
\textsuperscript{3} Massey University \hspace{0.4cm} \\
\textsuperscript{4}Ontario Tech University
}

\begin{document}

\maketitle
\begin{abstract}
Parameter-efficient fine-tuning (PEFT) methods are increasingly vital in adapting large-scale pre-trained language models for diverse tasks, offering a balance between adaptability and computational efficiency. They are important in Low-Resource Language (LRL) Neural Machine Translation (NMT) to enhance translation accuracy with minimal resources. However, their practical effectiveness varies significantly across different languages. We conducted comprehensive empirical experiments with varying LRL domains and sizes to evaluate the performance of 8 PEFT methods with in total of 15 architectures using the SacreBLEU score. We showed that 6 PEFT architectures outperform the baseline for both in-domain and out-domain tests and the Houlsby+Inversion adapter has the best performance overall, proving the effectiveness of PEFT methods.

\end{abstract}

\section{Introduction}
Advances in large-scale pre-trained language models have transformed the field for high-resource languages \cite{min2023recent}, but these data and compute-hungry models are not viable for the more-than-7000 low-resource languages (LRLs) in the world \cite{stap-araabi-2023-chatgpt, robinson-etal-2023-chatgpt, zhang-etal-2023-dont}.
Ideal for the limitations of LRLs, parameter-efficient fine-tuning (PEFT) methods \cite{Houlsby2019parameter, pfeiffer2020mad, hu2021lora} are designed to strategically update a small number of parameters within a pre-trained model to be more efficient and adaptable without retraining the entire model. Their architecture significantly saves computational resources and storage space while achieving results comparable to full fine-tuning in downstream tasks \cite{ruder-etal-2022-modular}. \citet{ustun2022does} examined the applicability of 4 PEFT methods specifically in the context of language translation. 
Moreover, it did not address truly LRLs \citep{ustun2022does}, nor did it incorporate variation in domains that would allow for an assessment of the models' generalization capabilities.

As a result, while the PEFT methods have shown potential in fine-tuning specific tasks, domains, and languages, the effectiveness of this collection of PEFT methods for LRL translation has not been systematically examined. In this paper, we explore the performance of different PEFT architectures in the LRL Neural Machine translation (NMT) by comparing in-domain and out-of-domain test results, as well as training times. We also investigate the effectiveness of PEFT methods in translating LRLs, focusing specifically on their architectures and performance across various datasets.


The contributions of our paper are 1) comprehensive experimentation of PEFT architectures to reveal the suitability of translating non-Latin scripts and LRL pairs; 2) an in-depth assessment of 15 PEFT architectures using 8 distinct methods to evaluate their effectiveness in LRL translation; and 3) a systematic exploration of experimental settings, including variations in dataset domains and sizes, aimed at enhancing model generalization capabilities.
As the field continues to advance rapidly, these PEFT guidelines provide practical recommendations for improving LRL translations, thus narrowing the language gap. 

\begin{figure}[!htp]
    \centering
    \includegraphics[width=\columnwidth]{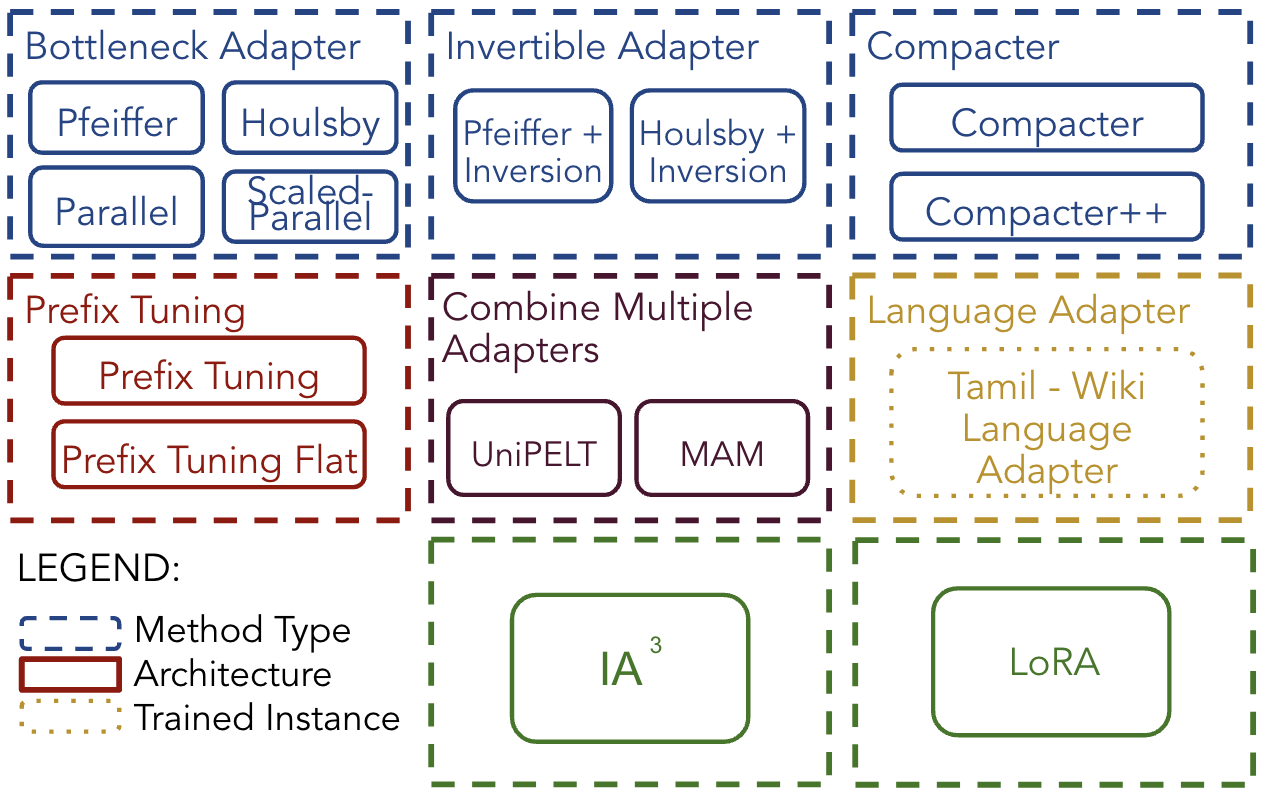}
    \caption{Full list of 8 PEFT methods and 15 architectures. Each color box represents a specific structure appearing in the PEFT methods. The same color represents the PEFT methods that share a similar structure.}
    \label{fig:peft}
\end{figure}
\section{The PEFT Methods}
We focus on the comparative performance of an extensive list of PEFT methods for LRL NMT under various settings (Figure~\ref{fig:peft}),  offering a broader and distinctive understanding of adapter utility.

Among all the PEFT methods, some share the same structure. For example, the bottleneck adapters include bottleneck feed-forward layers in each layer of a transformer model. These layers can be added to various positions within transformer blocks. The Houlsby adapter \citep{Houlsby2019parameter} adds the layers after both the multi-head attention and feed-forward blocks. The Pfeiffer adapter \citep{pfeiffer2020mad} only adds the layers after the feed-forward block. The Parallel adapter \citep{he2021towards} deploys the layers parallel to the transformer layers. Similarly, the invertible adapters share a similar architecture with bottleneck adapters but with an added invertible adapter layer to the language model embedding layer. The Compacter architecture replaces only the linear down-projection and up-projection with a parameterized hypercomplex multiplication layer \citep{karimi2021compacter}. 

In addition, Prefix Tuning is a lightweight alternative inspired by prompting \citep{li2021prefix} that introduces additional parameters in the multi-head attention blocks of each transformer layer. The LoRA method allows the training of specific dense layers in a neural network indirectly by optimizing the rank-decomposition matrices of specific dense layers during adaptation with the pre-trained weights frozen \citep{hu2021lora}. (IA)$^3$ is built to improve LoRA with modifications. While LoRA uses additive composition, (IA)$^3$ uses element-wise multiplication \citep{liu2022few}.

Some PEFT methods combine multiple methods. The Mix-and-Match (MAM) Adapter combines LoRA, Prefix Tuning, and Parallel adapter to form a new adapter \citep{he2021towards}. Similarly, UniPELT integrates bottleneck adapters, Prefix Tuning, and LoRA into a unified setup \citep{mao2021unipelt}. 

Lastly, the language adapter captures language-specific knowledge for application in various downstream tasks. It is not a distinct adapter architecture; rather, it represents a method of utilizing pre-existing architectures.  We expected that this approach would enhance the model's performance, given its preexisting familiarity with the language in question. We employed a pre-existing bottleneck adapter for diverse language datasets, training it with Masked Language Modelling on an extensive collection of articles \citep{pfeiffer20madx}.

\section{Experimental Setup}
\noindent\textbf{LRLs Selection } We chose Sinhala (SI), Tamil (TA), Hindi (HI), and Gujarati (GU) as our primary languages to run our translation task (See Table \ref{lang}). SI and TA were paired to run the translation task in both directions, and HI and GU were paired. 

\begin{table}[h]
\centering
\small
\resizebox{\linewidth}{!}{%
\begin{tabular}{@{}cccc@{}}
\toprule
\textbf{Language} &
  \textbf{Family} &
  \textbf{\begin{tabular}[c]{@{}c@{}}Joshi\\ class\end{tabular}} &
  \textbf{\begin{tabular}[c]{@{}c@{}}mBART coverage\\ in tokens (M)\end{tabular}} \\ \midrule
Hindi (HI)    & Indo Aryan & 4 & 1715 \\
Gujarati (GU) & Indo Aryan & 1 & 140  \\
Sinhala (SI)  & Indo Aryan & 1 & 243  \\
Tamil (TA)    & Dravidian  & 3 & 595  \\ \bottomrule
\end{tabular}%
}
\caption{\label{lang}Language details. The smaller the value of the ~\citet{joshi2020state} class, the more low-resource the language is. 
}
\label{lang_details}
\end{table}
\noindent\textbf{Data Collection  }  
The data summary is given in Table~\ref{dataset-summary}. More details about the datasets can be found in Appendix \ref{appendix:dataset}. Note that No Language Left Behind (NLLB) \citep{costa2022no} corpora are derived from metadata for bitext mining released by Meta AI, which lacks coverage and human quality control, and is only suitable for training purposes. Therefore, we performed an out-of-domain evaluation by using FLORES-101 \citep{goyal2022FLoRes} and FLORES-200 \citep{costa2022no} as the test dataset.

\begin{table*}[h]
    \centering
    \scriptsize 
    {%
    \begin{tabular}{c>{\centering\arraybackslash}p{8.3cm}ccccc}
    \toprule
    \textbf{Dataset} & \textbf{Quality} & \textbf{Languages} & \textbf{Train Size} & \textbf{Test Size} \\ \midrule
    FLORES-101 & Sourced from English Wikipedia and translated by professional translators & HI, GU, TA & Test only & 1k \\
    FLORES-200 & Sourced from web articles and translated by professional translators & SI & Test only & 1k \\
    NLLB & Automatically gathered from web sources and monolingual datasets, using web crawls and LASER3 encoders for parallel sentence identification & HI, GU, SI, TA & 25k, 100k & 2k \\
    Gvt & Parallel government documents dataset with manual cleaning and aligning & SI, TA & 25k & 2k \\
    Sam & Sourced both from existing corpora and new, diverse data collected via automated web crawling and sentence alignment, with human evaluation ensuring its reliability & HI, GU & 25k & 2k \\ \bottomrule
    \end{tabular}
    }
    \caption{\label{dataset-summary}Dataset statistics}
\end{table*}


\noindent\textbf{Pre-trained Model Selection} We performed baseline experiments by fine-tuning all parameters using several pre-trained models, including mBART-50 \citep{tang2020multilingual}, M2M-100 \citep{Fan2020Beyond}, and NLLB \citep{costa2022no}. Model selection process is given in Appendix~\ref{appendix:experiment}. Based on our results, we selected mBART-50 as our pre-trained model for the rest of our experiments. The mBART-50 model is a multilingual Sequence-to-Sequence (Seq2Seq) model. Its introduction aims to demonstrate the feasibility of developing multilingual translation models via the process of multilingual fine-tuning.


\noindent\textbf{Experimental Design}    
We also experimented by systematically varying the number of fine-tuned parameters that the method updates with the Houlsby adapter. It allowed us to investigate the impact of the number of parameters that are updated and then select the most suitable reduction factor for all the PEFT architectures.


The trainer employed in our study is 
sourced from the Adapter Transformers \citep{pfeiffer2020adapterhub}. Each adapter's performance was evaluated using the Sacre BiLingual Evaluation Understudy (SacreBLEU) Score \citep{post2018call}. Training details are given in Appendix \ref{appendix:experiment}.

We evaluated the performance of our PEFT architectures using direct fine-tuning with the pre-trained model as the baseline.  In total, we tested 15 PEFT architectures supported by the Hugging Face Adapter Hub \citep{pfeiffer2020adapterhub} trained on SI-TA 100k NLLB language dataset to identify the best methods for further analysis; both the NLLB test dataset and the FLoRes test dataset were used to test these models. We then narrowed down the selection to the top two methods with the highest SacreBLEU scores from each of the test results, the NLLB and the FLoRes dataset. An additional PEFT architecture was selected based on those that outperformed the baseline for both test datasets and with the shortest training time. 
Extensive experiments were then conducted with these top-selected methods across additional LRL and dataset sizes to determine the optimal configuration. After all experiments were completed, the average performance was calculated to mitigate any variation due to GPU randomness.

\section{Experimental Results}
\noindent 

\textbf{Number of fine-tuned parameters} Table~\ref{tab:performance-by-peft-parameters} shows that the performance of the same PEFT architecture can vary with the number of parameters. Initially, when the reduction factor is set to 2, both the in-domain and out-domain results show improvement over the baseline. Specifically, the in-domain performance increases significantly to \textbf{33.34}, while the out-domain performance also sees a modest improvement to 7.62. However, an interesting trend emerges as the reduction factor is further increased to 4: the in-domain test results begin to decline, dropping to 30.67, while the out-domain results experience only a slight increase (0.07 compared to reduction factor 2). This pattern suggests that increasing the reduction factor beyond 2 may lead to underfitting. Subsequent increases in the reduction factor exacerbate this trend, causing both in-domain and out-domain results to decrease compared to the reduction factor of 2. Therefore, the reduction factor of 2 is considered optimal for the remainder of the experiments, balancing model complexity with performance gains.

\begin{table}[h]
\centering
\resizebox{\linewidth}{!}{%
\begin{tabular}{@{}cccccc@{}}
\toprule
\textbf{\begin{tabular}[c]{@{}c@{}}Reduction\\ factor\end{tabular}} &
  \textbf{\begin{tabular}[c]{@{}c@{}}\# PEFT\\ parameters\end{tabular}} &
  \textbf{\begin{tabular}[c]{@{}c@{}}\% PEFT\\ parameters\end{tabular}} &
  \textbf{\begin{tabular}[c]{@{}c@{}}In-domain\\ (SacreBLEU)\end{tabular} } &
  \textbf{\begin{tabular}[c]{@{}c@{}} Out-domain\\ (SacreBLEU)\end{tabular}} &
  \textbf{\begin{tabular}[c]{@{}c@{}}Runtime \\ (hours)\end{tabular}} \\ \midrule
-  & -          & -    & 30.25          & 5.52          & 59.44          \\
2  & 50,405,376 & 7.62 & \textbf{33.34} & 7.62          & 78.65          \\
4  & 25,227,264 & 3.97 & 30.67          & \textbf{7.69} & \textbf{53.83} \\
8  & 12,638,208 & 2.03 & 31.05          & 7.52          & 93.58          \\
16 & 6,343,680  & 1.03 & 26.67          & 7.12          & 76.73          \\
32 & 3,196,416  & 0.52 & 23.81          & 7.35          & 74.18          \\ \bottomrule
\end{tabular}%
}
\caption{
Comparison of Fine-tuning Results with Different Reduction Factors on 100k NLLB SI-TA with mBART-50 Using the Houlsby Adapter. The first line represents the full fine-tuning baseline without using any PEFT architecture, the second line onwards shows fine-tuning with the Houlsby Adapter with different reduction factors. }
\label{tab:performance-by-peft-parameters}
\end{table}


\begin{table*}[h]
\resizebox{\linewidth}{!}{%
\begin{tabular}{ccccccc}
\toprule
\textbf{Architecture}          & \textbf{In-domain} & $\bm{\Delta\%}$           & \textbf{Out-domain} & $\bm{\Delta\%}$            & \textbf{Runtime (hours)}& $\bm{\Delta\%}$           \\ \midrule
Baseline        & 30.25          & -                 & 5.52            & -                 & 59.44           & -                \\
\textbf{Houlsby} & \underline{\textbf{33.34}}          & 10.20\% {[}1{]}   & \underline{\textbf{7.62}}            & 38.23\% {[}1{]}   & 78.65           & 32.32\% {[}10{]} \\
\textbf{Scaled-parallel}  & \underline{\textbf{33.04}} & 9.21\% {[}2{]}    & \textbf{6.62} & 20.00\% {[}7{]}   & 93.68 & 57.60\% {[}12{]} \\
\textbf{Pfeiffer+Inversion}    & \textbf{32.58}          & 7.69\% {[}3{]}    & \textbf{6.84}            & 24.06\% {[}4{]}   & 78.31           & 31.75\% {[}9{]}  \\
\textbf{MAM} & \textbf{33.26}          & 6.62\% {[}4{]}    & \textbf{6.51}            & 18.08\% {[}8{]}   & 95.73           & 61.05\% {[}13{]} \\
\textbf{Houlsby+Inversion}    & \textbf{32.23}         & 6.55\% {[}5{]}    & \underline{\textbf{7.42}}            & 34.51\% {[}2{]}   & 63.54           & 6.90\% {[}6{]}   \\
\textbf{Pfeiffer}        & \textbf{31.24}          & 3.27\% {[}6{]}    & \textbf{6.96}            & 26.25\% {[}3{]}   & \underline{\textbf{52.59} }         & -11.52\% {[}3{]} \\
Language Adapter (TA)              & 29.98          & -0.88\% {[}7{]}   & \textbf{6.31}            & 14.47\% {[}9{]}   & 98.23           & 65.26\% {[}14{]} \\
Parallel       & 27.63 & -8.66\% {[}8{]}   & \textbf{6.62} & 20.04\% {[}6{]}   & \textbf{26.85} & -54.83\% {[}1{]} \\
Prefix tuning  & 23.62          & -21.93\% {[}9{]}  & \textbf{6.71 }           & 21.72\% {[}5{]}   & 77.25           & 29.96\% {[}8{]}  \\
LoRA            & 18.63          & -38.41\% {[}10{]} & \textbf{5.76}            & 4.45\% {[}10{]}   & \textbf{58.1}            & -2.25\% {[}5{]}  \\
Compacter       & 13.36          & -55.82\% {[}11{]} & 4.27            & -22.61\% {[}11{]} & 106.56          & 79.27\% {[}15{]} \\
Compacter++     & 12.56          & -58.49\% {[}12{]} & 4.12            & -25.36\% {[}12{]} & 84.22           & 41.69\% {[}11{]} \\
Prefix tuning flat    & 12.25 & -59.50\% {[}13{]} & 3.93 & -28.75\% {[}13{]} & \textbf{55.29} & -6.98\% {[}4{]}  \\
(IA)$^3$            & 11.10          & -63.30\% {[}14{]} & 3.63            & -34.14\% {[}14{]} & 63.81           & 7.35\% {[}7{]}   \\
Unipelt         & 0.38           & -98.74\% {[}15{]} & 0.12            & -72.54\% {[}15{]} & \textbf{39.47 }          & -33.60\% {[}2{]} \\ \bottomrule
\end{tabular}
}
\caption{\label{tab:fullresults}
Full list of fine-tuning results with the 100k NLLB SI-TA language dataset. The table shows the predicted SacreBLEU score for both the In-domain test dataset (the NLLB test dataset), the Out-domain test dataset (the FLoRes test dataset), and the models' training time. $\Delta\%$ represents the percentage increase in terms of the baseline results. \textbf{Bold} means that the model's performance is better than the baseline (higher SacreBLEU score/shorter training time). \underline{Underline} means that the corresponding PEFT architectures are selected for further testing.} 
\end{table*}

\noindent\textbf{Top-4 Selected PEFT Architectures  }
To evaluate the PEFT architectures' performance, we compared their in-domain test, out-domain test, and training time for 100k NLLB SI-TA training dataset (Table~\ref{tab:fullresults}). For methods that did not surpass the baseline in both tests, we inferred that these methods are not suitable for tasks in LRL translation.

For NLLB in-domain testing, the Houlsby adapter performs the best at \underline{\textbf{33.34}} (10.20\% better than baseline), followed by Scaled-parallel (9.21\% improvement). For FLoRes out-of-domain testing, the Houlsby adapter remains the best at  \underline{\textbf{7.62}} (38.23\% better than baseline) followed by  Houlsby+Inversion adapter  (34.51\% improvement). The Pfeiffer adapter runs the fastest at \underline{\textbf{52.59}} while outperforming the baseline for both tests. 

\noindent\textbf{Domain Similarity of Test Dataset  }
We expanded our training to additional dataset domains (Appendix Table~\ref{all-test}). For the in-domain test, Houlsby adapter exhibits superior performance at \underline{\textbf{31.53}}; for the out-of-domain test, Houlsby+Inversion performs best at \underline{\textbf{10.02}} (a $0.1$ better than  Houlsby).
Since the FLoRes out-of-domain test results in a more robust and objective evaluation of the model's translation performance across many domains \citep{goyal2022FLoRes}, we prioritize the out-of-domain results and conclude that the Houlsby+Inversion adapter has the best performance overall. Lastly, in terms of training time (Appendix Table~\ref{runtime}), the Pfeiffer adapter has the shortest runtime as expected, saving 8 hours on average compared to the baseline.

\section{Discussion }

\begin{figure}[!htp]
    \centering
    \includegraphics[width=\columnwidth]{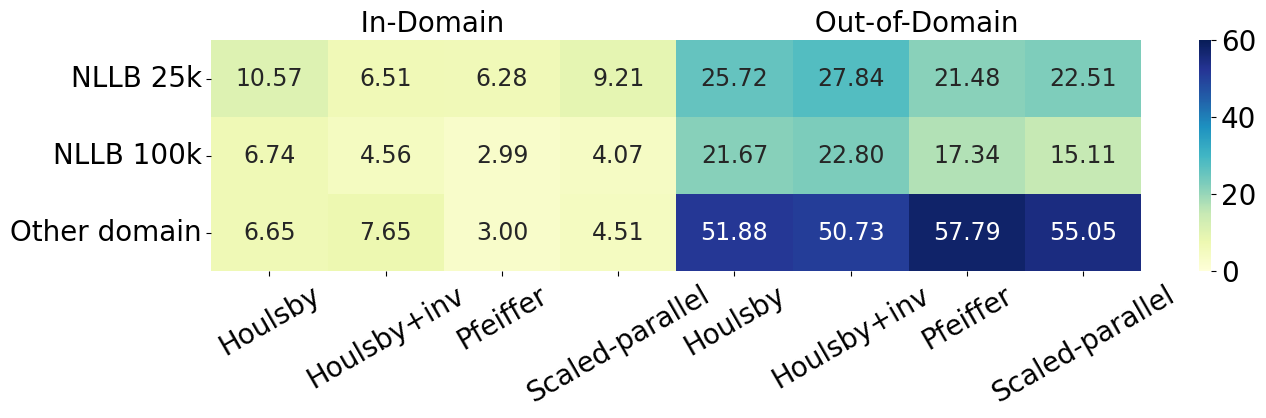}
    \caption{Average $\Delta\%$ compared to baseline for each dataset tested on in-domain and out-of-domain.}
    \label{fig:size}
\end{figure}

\noindent\textbf{Result Generalization  } 
Our results demonstrate the robust generalizability of our PEFT architectures across different training dataset sizes and domains. Figure~\ref{fig:size} shows that our model consistently outperforms the baseline, on average, in both in-domain and out-of-domain testing. 
Specifically for models trained on other domains, the $\Delta\%$ increase over the baseline is over 50\%, demonstrating the ability of PEFT methods to excel at tasks beyond their training domain. 
In terms of training dataset sizes, 
our selected PEFT architectures show a continuous trend for performance increase compared to the baseline. It is worth noting that our Table~\ref{all-test} in the appendix shows that increasing the training size has led to improved performance. 
However, the magnitude of the improvement difference shows diminishing returns, suggesting a potential saturation effect as identified in previous studies~\cite{lee-2021-improving-end}. 

\noindent\textbf{Effect of Bottleneck Architecture on LRL  } 
The similarity among the outperforming PEFT architectures highlighted in Table \ref{tab:fullresults} is that they all include bottleneck adapters in the architecture. While some other PEFT architectures such as UniPELT and   Compacter adapters  also adapt the bottleneck architecture, they do not exhibit comparable performance. In the subsequent discussion, we will examine the difference between these architectures and outperforming ones to find out which part of the bottleneck architecture design makes the performance better.

First, UniPELT integrates bottleneck adapters into a unified setup. Compared with the MAM adapter, UniPELT adds the bottleneck only after the Feed-Forward Network (FFN) layers \cite{mao2021unipelt}, while MAM also adds an adapter after the attention layer \cite{he2021towards}. 
This observation suggests that the efficiency of the architecture is not solely determined by the presence of the bottleneck adapter, but also by the specific placement of the adapter within the architecture. This could be because FFN learns task-specific text patterns \cite{geva-etal-2022-transformer}, while attention learns pairwise positional interactions. In our LRL translation task, pairwise positional interactions are more important than textual patterns. Removing the adapter from the attention layer may lead to a decrease in the SacreBLEU score.

Second, Compacter modifies the bottleneck adapter but does not achieve good results. The difference between a bottleneck adapter and Compacter is that it replaces the linear down and up projection with a parameterized hypercomplex multiplication (PHM) layer. This layer can break the importance of the parallel position in the translation task. This highlights the importance of the original up-and-down projection layer present in the bottleneck adapters.

\noindent\textbf{Adapter for Domain Adaptation  }
We found that in-domain testing performs better than out-of-domain testing due to memorizing patterns in the dataset, leading to falsely inflated performance. When fine-tuning on a new domain, rapid domain-specific overfitting and catastrophic forgetting reduce the performance on all other domains \cite{sennrich2015improving, barone2017regularization, bapna2019simple}. However, by freezing the parameters of the original pre-trained model and training only task-specific parameters, the adapter avoids catastrophic forgetting of the knowledge learned during pre-training and can maintain performance when testing in other domains \cite{mccloskey1989catastrophic, lai2022m, ustun2021multilingual}.

\begin{figure}[!htp]
    \centering
    \includegraphics[width=\columnwidth]{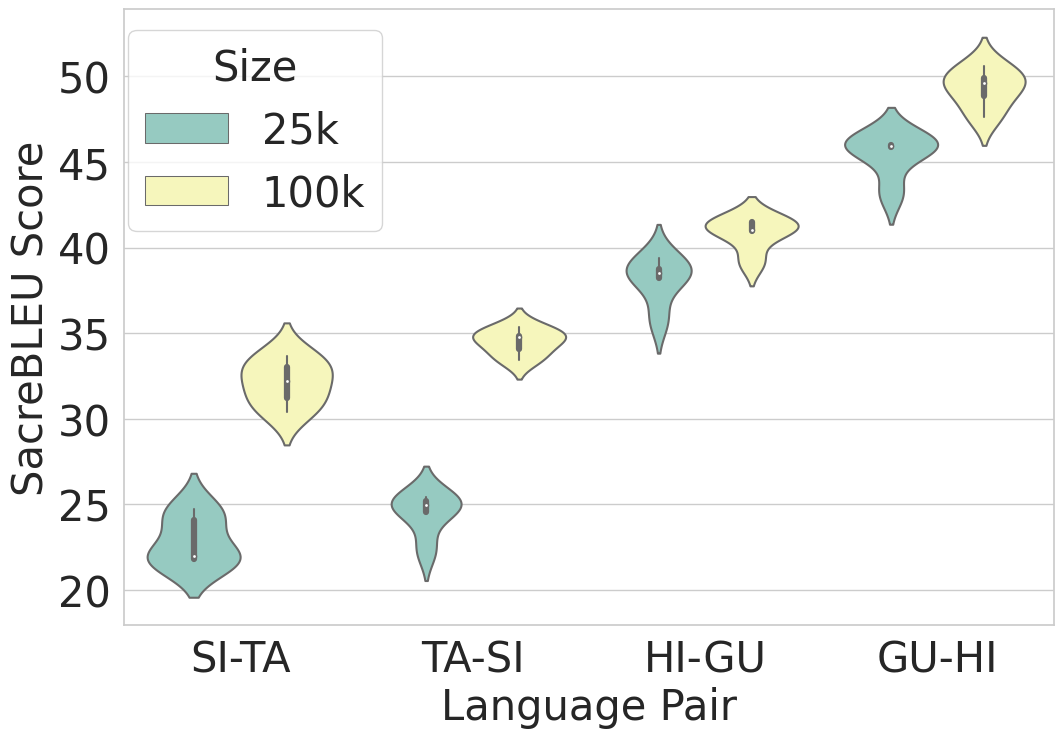}
    \caption{Performance of LRL Translation Pairs by Fine-Tuning Dataset Size (In-Domain only).}
    \label{fig:languages}
\end{figure}

\noindent\textbf{Language Family and Pre-Training Size} 
We observed notable disparities in performance among different language pairs (Figure~\ref{fig:languages}). The LRL SI-TA pair demonstrates lower performance with a smaller dataset size (i.e., 25k) but improves as the dataset size increases, suggesting that the amount of training data is a critical factor in enhancing the translation quality for LRL~\cite{lee2022pre}.

The SI-TA pair yields lower performance compared to the HI-GU pair, underscoring the intricate dynamics of linguistic relationships and the availability of resources (Table~\ref{lang_details}). Linguistically, HI, GU and SI are part of the Indo-Aryan language family, while TA is Dravidian; thus suggesting the lower performance of SI-TA. Notably, GU's closer linguistic affinity to HI may have facilitated enhancing its performance through cross-lingual transfer, despite its smaller pre-training dataset size.  However, its smaller gains due to dataset size increase may be due to the high-resource saturation of HI. 

\section{Conclusion}


Our study delved into a wide range of PEFT methods to identify the most effective ones for LRL-NMT. Particularly focusing on non-Latin scripts and LRL-to-LRL translation pairs, our research stands as a valuable guide for LRL-NMT. We found that certain adapters consistently outperformed others, offering enhanced translation accuracy and efficiency in challenging linguistic contexts.  Furthermore, the adapters' effectiveness was tested and generalized across various dataset domains and sizes, ensuring the applicability of our findings to a broad spectrum of LRL scenarios. 
Looking ahead, these insights pave the way for further advancements in PEFT methods, aiming to optimize the balance between efficiency and quality in NMT, especially in the challenging context of LRL.

\section*{Limitation and Future Work}
\noindent\textbf{Language Specific Adapters}   We tested the PEFT architectures at adapting to our LRLs, and not the specific fine-tuned models of language-specific adapters.  We hope this comparison can provide an agnostic baseline for others to follow.  Surprisingly, the language adapter we tested does not perform above the baseline; therefore, we need to explore other language-specific fine-tuning strategies.  In the future, we will explore more language-specific adapter; but the scope of this study only covers the generic PEFT architectures.  

\noindent\textbf{Increase Domain}    While it is worth noting that three of the four LRLs we have provided translations for belong to the Indo-Aryan language family and the other one is a Dravidian language, we suggest broadening our experimentation to include more diverse languages to increase the credibility of our results. As with dataset sizes of 100k and 25k, we could experiment with sizes in between.

\noindent\textbf{Evaluation Criteria} Our assessment of translation performance relied on SacreBLEU scores, but relying on a single metric may not be sufficient to support our conclusions. In future research to evaluate the model's performance, it would be advantageous to use metrics such as ChrF and COMET, which are reportedly better correlated with human judgments \citep{dixit2023indicmt}. Additionally, the variations between distinct methods lack strong indications. Consequently, statistical significance tests would be fundamental to further confirm the significance of the improvements.

\noindent\textbf{PEFT Composition} This paper focuses solely on the impact of a single PEFT architecture. However, there is an ongoing exploration into the potential of combining multiple methods as a composition. AdapterHub recently published a paper that expanded its support to include various composition methods, including stack, fuse, split, and average \citep{poth2023adapters}.


\bibliography{anthology,custom}
\bibstyle{acl_natbib}
\appendix
\section{Appendix}
\subsection{Supplementary Material on Datasets}
\label{appendix:dataset}

\noindent\textbf{No Language Left Behind (NLLB)} The NLLB \citep{costa2022no} corpus consists of translation training datasets for low-resource languages and is automatically created through the process of bitext mining.
We employed a selection process based on the LASER score, where we chose the top 100,000 and 25,000 translation pairs from the selected language pair for dataset size variation. However, NLLB lacks coverage and human quality control due to the noisy nature of the entire procedure and is only suitable for training purposes.

\noindent\textbf{Government corpus (Gvt)} The government document corpus \citep{fernando2020data} is a multi-way parallel corpus for Sinhala, Tamil, and English. It comprises a range of official Sri Lankan government documents, including annual and committee reports, content sourced from government websites, procurement-related documents, and legislative acts.

\noindent\textbf{Samanantar corpus (Sam)} The Samanantar corpus \citep{ramesh2023samanantar} is the largest publicly available Parallel Corpora Collection for 11 Indic Languages. The data is derived from two sources: existing databases and new data automatically collected through web crawling and sentence alignment techniques.

\noindent\textbf{FLORES} The FLORES dataset \citep{goyal2022FLoRes} is a multiway multilingual translation evaluation dataset. FLORES-101 is comprised of translations from 842 unique web articles, comprising a total of 3001 sentences. 
Because all translations are fully aligned, the resulting dataset allows for a more accurate assessment of model quality on the long tail of LRLs, including the evaluation of many-to-many multilingual translation systems. The professional rigor and reliability of the results are strengthened by using an out-of-domain evaluation of this type, resulting in a more robust and objective evaluation of the model's translation performance across many domains. FLORES-200 expands the language coverage to twice that of FLORES-101. We used FLORES-200 \citep{costa2022no} for Sinhala since it is not in FLORES-101, and dev-test split for both FLORES-101 and FLORES-200.

\subsection{Supplementary Material on Experimental Setup}
\label{appendix:experiment}
\noindent 
\textbf{Selection of Pre-trained Models}. We conducted experiments with several MT models, such as mBART-50 \citep{tang2020multilingual}, M2M-100 \citep{Fan2020Beyond} and NLLB \citep{costa2022no}. Specifically, we fine-tuned these models on the SI-TA 100k NLLB language dataset to identify the most effective methods for further analysis. Both the NLLB test dataset and the FLoRes test dataset were utilized to evaluate the performance of these models. Subsequently, we narrowed down the selection criteria to prioritize models with high SacreBLEU scores, low runtime, and lower computational resources. We found that nllb-200-distilled-1.3B, the largest model that we experimented with, has the best performance in both the in-domain and out-domain test sets (Table \ref{tab:baselines-different-models}). However, the superior performance of larger models comes with the caveat of increased hardware requirements, making them less accessible for practitioners, particularly when it comes to LRLs. In contrast, mBART-50 offers a robust alternative that does not demand additional computational resources, making it a practical choice for LRL applications. With these factors into account, we chose mBART-50 for our experiments.

\begin{table}[h]
\centering
\small
\resizebox{\linewidth}{!}{%
\begin{tabular}{@{}ccccc@{}}
\toprule
\textbf{Model}          & \textbf{\# of parameters} & \textbf{In-domain} & \textbf{Out-domain} & \textbf{Runtime (hours)} \\ \midrule
m2m100-418M           & 483,905,536   & 32.35 & 6.03 & 75.96          \\
mbart-large-50        & 610,879,488   & 30.25 & 5.52 & 59.44          \\
nllb-200-distilled-600M & 615,071,744   & 35.15 & 9.25 & \textbf{50.49} \\
m2m100-1.2B            & 1,239,470,080 & 32.22 & 6.22 & 187.91         \\
nllb-200-distilled-1.3B & 1,370,636,288             & \textbf{37.75}     & \textbf{10.30}      & 60.96                    \\ \bottomrule
\end{tabular}%
}
\caption{Baseline experiments with different pre-trained models with the 100k NLLB SI-TA language dataset.}
\label{tab:baselines-different-models}
\end{table}

\label{sec:appendix}

\begin{table*}[t]
\centering
\tiny
\setlength{\tabcolsep}{1.5pt} 
\resizebox{\linewidth}{!}{%
\begin{tabular}{lll|llllllllll}
\hline
\multirow{2}{*}{Language} &
  \multirow{2}{*}{Dataset} &
  \multirow{2}{*}{Size} &
  \multicolumn{2}{l}{No PEFT} &
  \multicolumn{2}{l}{Houlsby} &
  \multicolumn{2}{l}{Houlsby+inv} &
  \multicolumn{2}{l}{Pfeiffer} &
  \multicolumn{2}{l}{Scaled-parallel} \\
 &
   &
   &
  In-domain  &
  FLoRes &
  In-domain  &
  FLoRes  &
  In-domain  &
  FLoRes &
  In-domain  &
  FLoRes &
  In-domain &
  FLoRes  \\ \hline
\multirow{3}{*}{SI-TA} &
  \multirow{2}{*}{NLLB} &
  25k &
  21.8171 &
  \multicolumn{1}{l|}{3.9573} &
  \textbf{24.7268 (+2.9097)} &
  \multicolumn{1}{l|}{5.7709} &
  21.6649 &
  \multicolumn{1}{l|}{\textbf{5.8532 (+1.8959)}} &
  21.9808 &
  \multicolumn{1}{l|}{5.4773} &
  24.0997 &
  5.3101 \\
 &
   &
  100k &
  30.3961 &
  \multicolumn{1}{l|}{5.4352} &
  \textbf{33.6794 (+3.2833)} &
  \multicolumn{1}{l|}{\textbf{7.6977 (+2.2625)}} &
  32.2317 &
  \multicolumn{1}{l|}{7.4188} &
  31.2395 &
  \multicolumn{1}{l|}{6.9635} &
  33.0374 &
  6.6186 \\
 &
  Gvt &
  25k &
  21.2982 &
  \multicolumn{1}{l|}{1.3255} &
  21.0242 &
  \multicolumn{1}{l|}{2.2491} &
  \textbf{21.6247 (+0.3265)} &
  \multicolumn{1}{l|}{2.1965} &
  19.5961 &
  \multicolumn{1}{l|}{\textbf{2.347 (+1.0215)}} &
  20.5064 &
  2.0723 \\ \cline{1-3}
\multirow{3}{*}{TA-SI} &
  \multirow{2}{*}{NLLB} &
  25k &
  22.3512 &
  \multicolumn{1}{l|}{5.3989} &
  25.1825 &
  \multicolumn{1}{l|}{6.641} &
  \textbf{25.434 (+3.0828)} &
  \multicolumn{1}{l|}{\textbf{7.0094 (+1.6105)}} &
  24.5575 &
  \multicolumn{1}{l|}{6.1987} &
  24.9486 &
  6.4323 \\
 &
   &
  100k &
  34.0925 &
  \multicolumn{1}{l|}{7.1264} &
  \textbf{35.3707 (+1.2782)} &
  \multicolumn{1}{l|}{8.3163} &
  34.8269 &
  \multicolumn{1}{l|}{\textbf{8.6788 (+1.5524)}} &
  34.7869 &
  \multicolumn{1}{l|}{7.9525} &
  33.4139 &
  7.8196 \\
 &
  Gvt &
  25k &
  \textbf{31.9105} &
  \multicolumn{1}{l|}{2.4346} &
  31.7150 &
  \multicolumn{1}{l|}{3.2406} &
  31.7034 &
  \multicolumn{1}{l|}{3.259} &
  28.6959 &
  \multicolumn{1}{l|}{3.2433} &
  28.86 &
  \textbf{3.3824 (+0.9478)} \\ \cline{1-3}
\multirow{2}{*}{HI-GU} &
  \multirow{2}{*}{NLLB} &
  25k &
  35.8082 &
  \multicolumn{1}{l|}{11.2997} &
  \textbf{39.3775 (+3.5693)} &
  \multicolumn{1}{l|}{12.3927} &
  38.2209 &
  \multicolumn{1}{l|}{12.4318} &
  38.7203 &
  \multicolumn{1}{l|}{12.4832} &
  38.4944 &
  \textbf{12.807 (+1.5073)} \\
 &
   &
  100k &
  39.1754 &
  \multicolumn{1}{l|}{12.0767} &
  \textbf{41.5658 (+2.3904)} &
  \multicolumn{1}{l|}{14.2947} &
  41.4993 &
  \multicolumn{1}{l|}{\textbf{15.057 (+2.9803)}} &
  40.9938 &
  \multicolumn{1}{l|}{14.5054} &
  41.0432 &
  14.2797 \\
 &
  Sam &
  25k &
  11.1118 &
  \multicolumn{1}{l|}{5.2094} &
  12.6581 &
  \multicolumn{1}{l|}{9.5945} &
  12.6111 &
  \multicolumn{1}{l|}{9.0768} &
  12.7405 &
  \multicolumn{1}{l|}{\textbf{9.9535 (+4.7441)}} &
  \textbf{12.8279 (+1.7161)} &
  9.9509 \\ \cline{1-3}
\multirow{3}{*}{GU-HI} &
  \multirow{2}{*}{NLLB} &
  25k &
  43.2111 &
  \multicolumn{1}{l|}{13.9272} &
  45.9313 &
  \multicolumn{1}{l|}{\textbf{17.3196 (+3.3924)}} &
  45.8927 &
  \multicolumn{1}{l|}{17.2129} &
  45.9704 &
  \multicolumn{1}{l|}{17.0236} &
  \textbf{46.341 (+3.1299)} &
  17.1825 \\
 &
   &
  100k &
  47.6282 &
  \multicolumn{1}{l|}{17.5709} &
  \textbf{50.6256 (+2.9974)} &
  \multicolumn{1}{l|}{19.3265} &
  49.5878 &
  \multicolumn{1}{l|}{19.0191} &
  48.826 &
  \multicolumn{1}{l|}{19.2495} &
  49.9162 &
  \textbf{19.4532 (+1.8823)} \\
 &
  Sam &
  25k &
  14.3543 &
  \multicolumn{1}{l|}{10.0847} &
  16.4453 &
  \multicolumn{1}{l|}{12.1565} &
  \textbf{16.6844 (+6.5997)} &
  \multicolumn{1}{l|}{13.0219} &
  16.5667 &
  \multicolumn{1}{l|}{13.0903} &
  16.6316 &
  \textbf{13.5055 (+3.4208)} \\ \hline
\multicolumn{3}{l|}{Average} &
  29.4296 &
  \multicolumn{1}{l|}{7.9872} &
  \textbf{31.5252} &
  \multicolumn{1}{l|}{9.9167} &
  30.9985 &
  \multicolumn{1}{l|}{10.0196} &
  30.3895 &
  \multicolumn{1}{l|}{9.8740} &
  30.8434 & 9.9012
   \\ \hline
\end{tabular}
}
\caption{\label{all-test}
Comparison of Fine-Tuning Results for Selected PEFT Methods Across Various Language Datasets and Dataset Sizes on the in-domain Test Datasets and FLoRes Test Datasets. In-domain means that the test dataset comes from the same distribution as the training dataset. \textbf{Bold} score means that the SacreBLEU score is the highest among all listed fine-tuning experiments within the same dataset.  }
\end{table*}

\begin{table*}[t]
\centering
\scriptsize
\resizebox{\linewidth}{!}{%
\begin{tabular}{lll|lllll}
\hline
Language                & Dataset                 & Size & No PEFT           & Houlsby              & Houlsby+inv & Pfeiffer             & Scaled-parallel      \\ \hline
                        &                        & 25k  & \textbf{00-14:22:48} & 00-22:10:17          & 00-17:20:46 & 00-08:41:54          & 00-16:36:20          \\
                        & \multirow{-2}{*}{NLLB} & 100k & 02-23:47:07          & 03-12:06:21          & 02-15:32:23 & \textbf{02-04:35:37 (-19:11:30)} & 03-21:40:44          \\
\multirow{-3}{*}{SI-TA} & Gvt             & 25k  & 01-20:35:13          & 00-23:09:18          & 01-06:25:42 & \textbf{00-10:29:23 (-01-10:05:50)} & 00-18:55:42          \\ \cline{1-3}
                        &                        & 25k  & \textbf{00-09:51:53} & 00-19:04:15          & 01-06:57:42 & 00-21:56:10          & 00-21:12:29          \\
                        & \multirow{-2}{*}{NLLB} & 100k & 03-23:18:56          & 03-21:35:37          & 03-00:14:17 & 03-19:41:43          & \textbf{02-13:40:04 (-01-09:38:52)} \\
\multirow{-3}{*}{TA-SI} & Gvt             & 25k  & 02-01:01:03          & 01-14:06:04          & 02-02:11:33 & 00-20:36:14          & \textbf{00-10:26:10 (-01-14:34:53)} \\ \cline{1-3}
                        &                        & 25k  & 00-07:42:33          & 00-17:45:38          & 00-10:43:29 & 00-15:47:21          & \textbf{00-06:50:13} \\
                        & \multirow{-2}{*}{NLLB} & 100k & {\color[HTML]{202124} 01-05:37:21} & 01-02:22:44          & 01-00:18:21 & \textbf{00-19:28:39 (-10:08:42)} & 00-22:16:01 \\
\multirow{-3}{*}{HI-GU} & Sam            & 25k  & 00-16:27:37          & 00-07:43:53          & 00-07:27:22 & 00-05:51:51          & \textbf{00-05:29:49 (-10:57:48)} \\ \cline{1-3}
                        &                        & 25k  & 00-07:34:30          & \textbf{00-04:59:17 (-02:35:13)} & 00-07:20:46 & 00-05:47:51          & 00-06:23:02          \\
                        & \multirow{-2}{*}{NLLB} & 100k & \textbf{00-20:17:54} & 01-07:19:39          & 01-02:59:59 & 00-21:23:38          & 00-20:35:02          \\
\multirow{-3}{*}{GU-HI} & Sam             & 25k  & {\color[HTML]{202124} 00-04:54:57} & 00-04:54:34) & 00-05:51:34 & 00-04:59:19          & \textbf{00-04:46:03 (-00:08:54)} \\ \hline
\multicolumn{3}{l|}{Average}                             & 01-06:57:39          & 01-07:06:28          & 01-04:57:00 & \textbf{00-23:16:38} & 01-00:04:18          \\ \hline
\end{tabular}
}
\caption{\label{runtime}
Comparison of Training Time for Selected PEFT Methods Across Language Datasets and Dataset Sizes. \textbf{Bold} time means that the training time is the shortest among listed fine-tuning experiments with the same dataset. }
\end{table*}
\begin{table}[h]
\centering
\small
\begin{tabular}{cc}
\toprule
\textbf{Parameter}           & \textbf{Value}           \\ \midrule
Evaluation Strategy      & Epoch           \\
Number of Training Epoch & 40              \\
Patience                 & 3               \\
Batch Size               & 2               \\
Metric for Best Model    & Evaluation SacreBLEU \\ \bottomrule
\end{tabular}
\caption{\label{trainer-parameter}
Full list of trainer parameters used and corresponding value.}
\end{table}
\noindent\textbf{Choice of Trainer  }
The integration of PEFT methods into language models is facilitated by a modification of AdapterHub, a centralized store of pre-trained adapter modules. 

In the context of language translation, the process involves utilizing a translation code to refine the pre-existing model and assess the performance of transformers to translation-oriented assignments. In this case, we used Seq2SeqTrainingArguments.

\noindent\textbf{GPU Details}
 It consists of Dell nodes, each equipped with four NVIDIA V100-32GB GPUs, 32 CPU cores, 32GB of GPU memory, and two Intel Silver 4216 Cascade Lake processors running at 2.1GHz. All GPUs are connected via NVLink and SXM2. They are well suited for processing large language models with a 7.0 capability. 
 
\noindent\textbf{Trainer Setup}
There are several parameters that we have specified for the execution of the model. For the evaluation strategy, the evaluation is done at the end of each epoch \citep{wolf-etal-2020-transformers}. We set the number of training epochs to 40 so that the model could be finished running in a maximum of 4 days.  The patience level is set to 3 based on some small experiments. A lower level of patience will cause the model to stop too early as there is still room for improvement; a higher level of patience will cause overfitting, and the model will only stop until the last epoch; there will be no early stopping, which is not what we expected. Since our task is simple fine-tuning, we set the batch size to 2. A smaller batch size introduces more stochasticity into the training process by updating the model parameters more frequently.

\noindent\textbf{Evaluation Metrics}
SacreBLEU \citep{post2018call} offers benefits over BLEU scores, which cannot be directly compared across papers, as it allows for easy computation of shareable, comparable, and reproducible SacreBLEU scores.

\section{Direct Fine-Tuning Results With Selected PEFT Architectures Across Different Domains}
\label{appendix:all}
We selected Houlsby, Houlsby+Inversion, and Scaled-Parallel Adapter for the next experiments based on their performance, with Houlsby emerging as the best performer for both testing results. Pfeiffer adapter was selected for its short training time compared to the baseline. The results displayed in Table~\ref{all-test} indicate that the Houlsby adapter exhibit superior performance over all other methods in the in-domain test with an average SacreBLEU score of \textbf{31.5252}. For the FLoRes test dataset, Houlsby+Inversion performs better with an average SacreBLEU score of \textbf{10.0196}, a 0.1 difference from Houlsby.

In terms of training time shown in~\ref{runtime}, the Houlsby adapter does not have the advantage and even becomes the longest runtime on average. The Pfeiffer adapter, which we chose for its runtime, has the shortest runtime as expected, saving 8 hours on average compared to the baseline.

\end{document}